\documentclass{article}
\usepackage{microtype}
\usepackage{graphicx}
\usepackage{xcolor}
\usepackage{subfigure}
\usepackage{booktabs} 
\usepackage{amsfonts,amsmath,amssymb,amsthm,mathtools}       
\usepackage{enumitem}
\usepackage{tabularx}
\usepackage{algorithm}
\usepackage{algorithmic}
\usepackage{nicefrac}

\usepackage{tikz}
\usetikzlibrary{shapes.geometric}
\usepackage{pgfplots}
\pgfplotsset{compat=newest}

\definecolor{yellowgreen}{rgb}{0.6039215686274509, 0.803921568627451, 0.19607843137254902}
\definecolor{cH}{HTML}{001C7F}
\definecolor{cX}{HTML}{8C0900}

\graphicspath{{fig/}}

\usepackage{hyperref}

\usepackage[accepted]{icml2021arxiv}

\newcommand{\gp}{\textsc{gp}}
\newcommand{\cg}{\textsc{cg}}
\newcommand{\hmc}{\textsc{hmc}}
\newcommand{\gpghmc}{\textsc{gpg-hmc}}

\newcommand{\trans}{^{\top}}
\newcommand{\Trans}{^{\top}}
\newcommand{\R}{\mathbb{R}}
\newcommand{\vect}[1]{\operatorname{vec}(#1)}
\renewcommand{\v}[1]{\boldsymbol{#1}}

\newcommand{\op}[1]{\operatorname{#1}}

\newcommand\thefont{\expandafter\string\the\font}

\newtheorem{definition}{Definition}

\icmltitlerunning{High-Dimensional Gaussian Process Inference with Derivatives}

\begin{document}

\twocolumn[
\icmltitle{High-Dimensional Gaussian Process Inference with Derivatives}

\begin{icmlauthorlist}
\icmlauthor{Filip de Roos}{tue,mpiis}
\icmlauthor{Alexandra Gessner}{tue,mpiis}
\icmlauthor{Philipp Hennig}{tue,mpiis}
\end{icmlauthorlist}

\icmlaffiliation{tue}{Department of Computer Science, University of T{\"u}bingen, T{\"u}bingen, Germany}
\icmlaffiliation{mpiis}{Max Planck Institute for Intelligent Systems, T{\"u}bingen, Germany}

\icmlcorrespondingauthor{Filip de Roos}{filip.de.roos@tuebingen.mpg.de}


\vskip 0.3in
]

\printAffiliationsAndNotice{}  

\begin{abstract}
Although it is widely known that Gaussian processes can be conditioned on observations of the gradient, this functionality is of limited use due to the prohibitive computational cost of $\mathcal{O}(N^3 D^3)$ in data points $N$ and dimension $D$.
The dilemma of gradient observations is that a single one of them comes at the same cost as $D$ independent function evaluations, so the latter are often preferred.
Careful scrutiny reveals, however, that derivative observations give rise to highly structured kernel Gram matrices for very general classes of kernels (inter alia, stationary kernels).
We show that in the \emph{low-data} regime $N<D$, the Gram matrix can be decomposed in a manner that reduces the cost of inference to $\mathcal{O}(N^2D + (N^2)^3)$ (i.e.,~linear in the number of dimensions) and, in special cases, to $\mathcal{O}(N^2D + N^3)$.
This reduction in complexity opens up new use-cases for inference with gradients especially in the high-dimensional regime, where the information-to-cost ratio of gradient observations significantly increases.
We demonstrate this potential in a variety of tasks relevant for machine learning, such as optimization and Hamiltonian Monte Carlo with predictive gradients.
\end{abstract}

\section{Introduction}
\label{sec:introduction}

The closure of Gaussian processes (\gp s) under linear operations is well-established \citep[Ch. 9.4]{rasmussen2006gaussian}.
Given a Gaussian process $f\sim\mathcal{GP}(\mu, k)$, with mean and covariance function $\mu$ and $k$, respectively, a linear operator $\mathcal{L}$ acting on $f$ induces another Gaussian process $\mathcal{L} f \sim \mathcal{GP}(\mathcal{L}\mu, \mathcal{L}k\mathcal{L}')$ for the operator $\mathcal{L}$ and its adjoint $\mathcal{L}'$.
The linearity of \gp s has found extensive use both for conditioning on projected data, and to perform inference on linear transformations of $f$.
Differentiation is a linear operation and thus, gradient information has found considerable attention in a wide variety of applications that use \gp~models.
However, each gradient observation of $\nabla f \in \mathbb{R}^D$ induces a block Gram matrix $\nabla k \nabla' \in \mathbb{R}^{D\times D}$.
As dimension $D$ and number of observations $N$ grow, inference with gradient information quickly becomes prohibitive with the na\"ive scaling of $\mathcal{O}(N^3 D^3)$.
In other words, one gradient observation comes at the same computational cost as $D$ independent function evaluations and thus becomes increasingly disadvantageous as dimensionality grows.
This is not surprising, as the gradient contains $D$ elements and thus bears information about every coordinate.
The unfavorable scaling has confined \gp~inference with derivatives to low-dimensional settings in which the information gained from gradients outweighs the computational overhead (cf.~Sec.~\ref{sec:related_work} for a review).

In this work we show that gradient Gram matrices possess structure that enables inversion at cost linear in $D$.
This discovery unlocks the previously prohibitive use of gradient evaluations in high dimensional spaces for nonparametric models.
Numerous machine learning algorithms that operate on high-dimensional spaces are guided by gradient information and bear the potential to benefit from an inference mechanism that avoids discarding readily available information.
Examples for such applications that we consider in this work comprise optimization, linear algebra, and gradient-informed Markov chain Monte Carlo.

\paragraph{Contributions} 
We analyze the structure of the Gram matrix with derivative observations for stationary and dot product kernels and report the following discoveries:

\begin{itemize}	
	\item The Gram matrix can be decomposed to allow \emph{exact} inference in $\mathcal{O}(N^2 D + (N^2)^3)$ floating point operations, which is useful in the limit of few observations ($N<D$).
	\item We introduce an efficient approximate inference scheme to include gradient observations in \gp s even as the number of high-dimensional observations increases. It relies on exact matrix-vector products (MVP) and an iterative solver to approximately invert the Gram matrix. This implicit MVP avoids constructing the whole Gram matrix and thereby reduces the memory requirements from $\mathcal{O}((ND)^2)$ to $\mathcal{O}(N^2 + ND)$.
	\item We demonstrate the applicability of the improved scaling in the low-data regime on high-dimensional applications ranging from optimization to Hamiltonian Monte Carlo.
	\item We explore a special case of inference with application to probabilistic linear algebra for which the cost of inference can be further reduced to $\mathcal{O}(N^2 D + N^3)$.
\end{itemize}

\section{Theory} 
\label{sec:theory}

\begin{figure*}[ht]
\centering
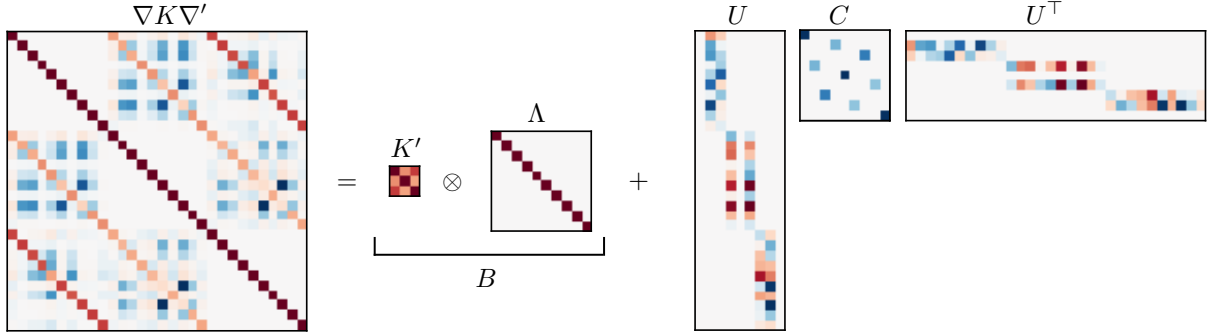
\caption{Gram matrix built from three 10-dimensional gradient observations using a stationary isotropic exponential quadratic kernel. Explicit expression \emph{(left)} and its decomposition into a Kronecker product $B$ and low-rank correction $UCU^T$ that allows for efficient inversion using Woodbury's matrix lemma (cf. Sec.~\ref{sub:implementation}) if $N<D$ \emph{(right)}. Positive values are colored red, negative blue and white indicates 0.}
\label{fig:gradient_kernel}
\end{figure*}

Kernel matrices of Gaussian processes (\gp s) built from gradient observations are highly structured.
In this section, after reviewing \gp s, we show that for standard kernels, the kernel Gram matrix can be decomposed into a Kronecker product with an additive low-rank correction, as exemplified in Fig.~\ref{fig:gradient_kernel}.
Exploiting this structure, \emph{exact} \gp~inference with gradients is feasible in $\mathcal{O}(N^2 D + (N^2)^3)$ operations instead of $\mathcal{O}((DN)^3)$ when inverting the kernel matrix exactly.
Furthermore, the same structure enables storage of $\mathcal{O}(N^2 + ND)$ values instead of $\mathcal{O}((ND)^2)$.

\subsection{Gaussian Processes} 

\label{sub:gaussian_processes}

\begin{definition}
 A Gaussian process $f\sim \mathcal{GP}(\mu, k)$ is a random process with mean function $\mu:\R^D \mapsto \R$ and covariance function $k: \R^D\times \R^D \mapsto \R$ such that $f$ evaluated at a finite set of inputs follow a multi-variate normal distribution \citep[Ch. 2.2]{rasmussen2006gaussian}. 
\end{definition}

\gp s are popular nonparametric models with numerous favorable properties, of which we highlight their closure under linear operations.
A linear operator acting on a \gp~results again in a \gp.
Let $\mathcal{L}, \mathcal{M}$ be linear operators acting on $f$.
Then the joint distribution of $\mathcal{L} f$ and $\mathcal{M}f$ is:
\begin{equation}
  \begin{bmatrix}
  \mathcal{L} f\\
  \mathcal{M} f
  \end{bmatrix}
  \sim 
  \mathcal{GP}\left( 
  \begin{bmatrix}
    \mathcal{L} \mu\\
    \mathcal{M} \mu
  \end{bmatrix},
  \begin{bmatrix}
    \mathcal{L} k \mathcal{L}' & \mathcal{L} k \mathcal{M}'\\           
    \mathcal{M} k \mathcal{L}' & \mathcal{M} k \mathcal{M}' \\
  \end{bmatrix}
    \right), 
    \label{eq:lin_gp}
\end{equation}
where $\mathcal{L}'$ and $\mathcal{M}'$ act on the second argument of the covariance function $k$.
The conditional $\mathcal{L} f \mid \mathcal{M} f$ is obtained with standard Gaussian algebra and requires the inversion of $\mathcal{M} k \mathcal{M}'$.\\
Examples of linear operators comprise projections, integration, and differentiation.
We focus here on inference on either $f$ itself, its gradient $\v{g} = \nabla f$, or its Hessian matrix $H = \nabla\nabla\Trans f$ conditioned on gradient observations, i.e.~$\mathcal{L} = \{\operatorname{Id}, \nabla, \nabla\nabla\Trans\}$ and $\mathcal{M} = \nabla$.

\paragraph{Notation}
We collect gradient observations $\v{g}_a \in \R^D$ at locations $\v{x}_a \in \R^D$, $a=1,\dots,N$ which we vertically stack into the data matrices $X \in \R^{D\times N}$ and $G \in \R^{D\times N}$.
The object of interest is the Gram matrix $\nabla K \nabla' \in \R^{DN\times DN}$ where $K = k(X, X)$ and $\nabla, \nabla'$ act w.r.t. all elements of $X$.
We let subscripts $a, b$ identify indices related to \emph{data points}, e.g., $\v{x}_a, \v{x}_b$. Superscript indices $i, j$ refer to indices along the input \emph{dimension}.
In further abuse of notation we will let the operation $\tilde{X} = X-\v{c}$ denote the subtraction of $\v{c}$ from each column in $X$.

\subsection{Exploiting Kernel Structure} 

The efficient inversion of $\nabla K \nabla'$ relies on its somewhat repetitive structure involving a Kronecker product (Fig.~\ref{fig:gradient_kernel}) caused by application of the product and chain rule of differentiation to the kernel. The Kronecker product $A\otimes B$ produces a matrix with blocks $a_{ij} B$ (cf.~\citet{vanloan2000kronecker} and 
Appendix~\ref{A:la} 
for properties of the Kronecker product).\\
Any kernel $k(\v{x}_a, \v{x}_b)$ with inputs $\v{x}_a, \v{x}_b \in \R^D$, can be equivalently written on terms of a scalar function $r:\R^D\times \R^D\mapsto\R$ as $k\left(r(\v{x}_a, \v{x}_b)\right) \eqqcolon k_{ab} (r)$.
Note the general definition of $r$, which in particular is more general than stationarity.
Since $k$ is also a scalar function of $\v{x}_a$ and $\v{x}_b$, $r$ could be equal to $k$ if there is no way to further condense the relationship between $\v{x}_a$ and $\v{x}_b$.

\begin{definition}
 Write $k(\v{x}_a, \v{x}_b) = k_{ab} (r)$ with $r\in\R$. Define $k_{ab}' = \frac{\partial k(\v{x}_a, \v{x}_b)}{\partial r}$, $k_{ab}'' = \frac{\partial^2 k(\v{x}_a, \v{x}_b)}{\partial r^2}$ and ${\partial_a}^i = \frac{\partial}{{\partial \v{x}_a}^i}$ and similarly for ${\partial_b}^j$.
 The derivatives of $k$ w.r.t. ${\partial \v{x}_a}^i$ and ${\partial \v{x}_b}^j$ can be written as
  \begin{align}
    {\partial_a}^i k_{ab}(r) &= k_{ab}'(r)\, {\partial_a}^i r \nonumber \\
    {\partial_b}^j k_{ab}(r) &= k_{ab}'(r)\, {\partial_b}^j r \nonumber\\ 
    {\partial_a}^i {\partial_b}^j k(r) &= k_{ab}'(r)\cdot {\partial_a}^i {\partial_b}^j r  + k_{ab}''(r)\cdot ({\partial_a}^i r) ({\partial_b}^j r)
    \label{eq:kernel_general}
  \end{align}
\end{definition}
This expression is still general but we can already see that the derivatives of $k$ w.r.t. $r$ depend only on the indices $a, b$ of the data points and form $N\times N$ matrices that we call $K'$ and $K''$.
Importantly, they do \emph{not} on dimensional indices $i,j$.
While the abundance of indices invites for a tensor-like implementation, the prime gain comes from writing Eq.~\eqref{eq:kernel_general} in matrix form.
Doing so permits linear algebra operations that are not applicable to tensors.
We specify the matrix form of the general expression of Eq.~\eqref{eq:kernel_general} for two overarching classes of kernels: the dot product and stationary class of kernels.
For these kernels, $r$ is defined as
\begin{align*}
 r &=(\v{x}_a- \v{c})\trans\Lambda(\v{x}_b - \v{c})\quad \text{\emph{(dot product kernels)}},\\
 r &=(\v{x}_a - \v{x}_b)\trans\Lambda(\v{x}_a - \v{x}_b)\quad \text{\emph{(stationary kernels)}},
\end{align*}
with an arbitrary offset $\v{c}$ and a symmetric positive definite scaling matrix $\Lambda$.

\paragraph{Dot Product Kernels}
For dot product kernels the Gram matrix of the gradients in Eq.~\eqref{eq:kernel_general} is
\begin{equation*}
  {\partial_a}^i {\partial_b}^j k(r) = \\
  k_{ab}'(r)\cdot \Lambda^{ij}  + k_{ab}''(r)\cdot [\Lambda (\v{x}_b-\v{c})]^{i} [\Lambda (\v{x}_a-\v{c})]^{j}, 
\end{equation*}
which can be written as a low-rank update to a Kronecker products as (see 
Appendix~\ref{A:dot_product} 
for the derivation)
\begin{equation}
  K'\otimes \Lambda  + (I\otimes \Lambda(X-\v{c}))\, C\, (I\otimes (X-\v{c})\trans \Lambda),
  \label{eq:dot_kron}
\end{equation}
where $U = I\otimes \Lambda(X-\v{c})$ is of size $DN\times N^2$.
The matrix $C\in \R^{N^2\times N^2}$ is a permutation of $\operatorname{diag} \left(\vect{K''}\right)$,  i.e.~a diagonal matrix that has the elements of $K''$ on its diagonal, such that $C\vect{M}=\vect{K''\odot M\trans}$, for $M\in\R^{N\times N}$. Here, $\odot$ denotes the Hadamard (element-wise) product.

\paragraph{Stationary Kernels}
The gradient structure of Eq.~\eqref{eq:kernel_general} for stationary kernels looks similar to the dot product case: 
\begin{multline}
{\partial_a}^i {\partial_b}^j k(r) =\\ -k_{ab}'(r)\cdot \Lambda^{ij}  - k_{ab}''(r)\cdot [\Lambda(\v{x}_a - \v{x}_b)]^{i}  [\Lambda(\v{x}_a - \v{x}_b)]^{j}.
\end{multline}
Similarly to the dot product kernel, this also takes the matrix form
\begin{equation}
 K'\otimes \Lambda  + U\, C\, U\trans,
\end{equation}
where the first term $K'\otimes \Lambda$ as well as $C$ remain unaltered.
Compared to Eq.~\eqref{eq:dot_kron}, however, the expression of $U$ changes and appears more intricate due to the interchange of subscripts (see 
Appendix~\ref{A:stationary}
for more details).
It can be written as $U=(I \otimes \Lambda X) L$ with a sparse $N^2 \times N^2$ matrix $L$ that substracts $\Lambda \v{x}_a$ from all columns of the $a^\text{th}$ block of the block diagonal matrix $I \otimes \Lambda X$.
Figure \ref{fig:gradient_kernel} illustrates the decomposition for the exponential quadratic a.k.a.~radial basis function (RBF) kernel.
Explicit expressions for a few common dot product and stationary kernels are found in 
Appendix~\ref{A:dotprod_examples} and \ref{A:stationary_examples}.

\subsection{Implementation} 
\label{sub:implementation}
The structure of the Gram matrix promotes two important tricks that enable efficient inference with gradients.
Computational gains come into play when $N<D$, but for any choice of $N$, the uncovered structure enables massive savings in storage and enables efficient approximate inversion schemes.

\paragraph*{Low-Data Regime}
In the high-dimensional regime with a small number of observations $N<D$ the inverse of the Gram matrix can be efficiently obtained from Woodbury's matrix inversion lemma \citep{woodbury1950inverting}
\begin{multline}
  \left( B+ UCU\trans \right)^{-1} =\\
   B^{-1} - B^{-1}U\left( C^{-1} + U\trans B^{-1}U \right)^{-1}U\trans B^{-1}
  \label{eq:woodbury}
\end{multline}
(if the necessary inverses exist), combined with inversion properties of the Kronecker product.
If $B$ is cheap to invert and the dimensions of $C$ are smaller than the ones of $B$, then the above expression can drastically reduce the computational cost of inversion. In our case $B=K'\otimes\Lambda$ for which the inverse $B^{-1}=(K')^{-1}\otimes \Lambda^{-1}$ requires the inverse of the $N\times N$ matrix $K'$.
The main bottleneck is the inversion of the $N^2\times N^2$ matrix $C^{-1} + U\trans B^{-1}U$ which requires $\mathcal{O}(N^6)$ operations, which is still a benefit over the na\"ive scaling when $N<D$.
The low-rank structure along with properties of Kronecker products leads to a general solution of the linear system $[\nabla K \nabla']\vect{Z}=\vect{G}$ of the form
\begin{equation}
  Z = \Lambda^{-1} G (K')^{-1} - \tilde{X} Q
  \label{eq:solution}
\end{equation}
for dot product kernels with $\tilde{X} = X - \v{c}$ and gradient observations $G$.
$Q$ is the unvectorized solution to
\begin{equation}
  (C^{-1} + U\trans B^{-1}U) \vect{Q}=\vect{\tilde{X}\trans G(K')^{-1}}.
  \label{eq:inner}
\end{equation}
Stationary kernels give rise to a similar expression (cf.~Appendix~\ref{A:woodbury_smallN}).

\paragraph*{General Improvements}
The cubic computational scaling is frequently cited as the main limitation of \gp~inference, but often the quadratic storage is the real bottleneck. For gradient inference that is particularly true due to the required $\mathcal{O}((ND)^2)$ memory.
A second observation that arises from the decomposition is that the the Gram matrix $\nabla K \nabla'$ is fully defined by the much smaller matrices $K'$, $K''$ (both $N\times N$), $\Lambda X$ ($D\times N$) and $\Lambda$ ($D\times D$, but commonly chosen diagonal or even scalar).
Thus, it is sufficient to keep only those in memory instead of building the whole $DN\times DN$ matrix $\nabla K \nabla'$, which requires at most $\mathcal{O}(N^2 + ND + D^2)$ of storage.
Importantly, this benefit arises for $D>1$ and for any choice of $N$.
It is further known how these components act on a matrix of size $D\times N$.
For dot product kernels, a multiplication of the Gram matrix with vectorized matrix $V\in\R^{D\times N}$ is obtained by
\begin{equation}
  (\nabla K \nabla') \vect{V}= \Lambda V K' + \Lambda \tilde{X} (K''\odot V\trans \Lambda \tilde{X})
  \label{eq:matvec}
\end{equation}
A similar expression is obtained for stationary kernels, see Appendix~\ref{A:woodbury_general}.
This routine can be used with an iterative linear solver \cite{gibbs1997cggp,gardner2018gpytorch} to exactly solve a linear system in $DN$ iterations or to obtain an approximate solution in even fewer iterations. The multiplication routine is further amenable to preconditioning which can drastically reduces the required number of iterations \cite{eriksson2018scaling}, as well as popular sparsification techniques used to lower the computational cost.

\section{Related Work} 
\label{sec:related_work}

Exact derivative observations have been used to condition \gp s on linearizations of dynamic systems \cite{solak2003derivative} as a way to condense information in dense input regions. This required the number of replaced observations to be larger than the input dimension in order to benefit.
Derivatives have also been employed to speed up sampling algorithms by querying a surrogate model for gradients \cite{rasmussen2003hybridmc}. In both previous cases the algorithms were restricted to low-dimensional input but showed improvements over baselines despite the computational burden. In Sec.~\ref{sub:hmc} we will revisit the idea of sampling in light of our results.

Modern \gp~models that use gradients always had to rely on various approximations to keep inference tractable.
\citet{solin2018modeling} linearly constrained a \gp~to explicitly model curl-free magnetic fields \cite{jidling2017linearly}. This involved using the differentiation operator and was made computationally feasible with a reduced rank eigenfunction expansion \cite{solin2020hilbert}. 
\citet{angelis2020sleipnir} extended the quadrature Fourier feature expansion (QFF) \cite{mutny2018QFF} to derivative information. The authors used it to construct a low-rank approximation for efficient inference of ODEs with a high number of observations. Derivatives have also been included in Bayesian optimization but mainly in low-dimensional spaces \cite{osborne2009bo,lizotte2008practicalbo}, or by relying on a single gradient observation in each iteration \cite{wu2017bayesian}.

A more task-agnostic approach was presented by \citet{eriksson2018scaling}. The authors derived the gradient Gram matrix for the structured kernel interpolation (SKI) approximation \cite{wilson2015kernel}, and its extension to products SKIP \cite{gardner2018product}. This was used in conjunction with fast matrix-vector multiplication on GPUs \cite{gardner2018gpytorch} and a subspace discovery algorithm to make inference efficient. \citet{tej2020DBQPG} used  a similar approach but further incorporated Bayesian quadrature with gradient inference to infer a noisy policy gradient for reinforcement learning to speed up training.


An obvious application of gradients for inference is in optimization and in some cases linear algebra. These are two fields we will discuss further in Sec.~\ref{sec:applications}.
Probabilistic versions of linear algebra and quasi-Newton algorithms can be constructed by modeling the Hessian with a matrix-variate normal distribution and update the belief from gradient observations \cite{hennig2015probabilistic,wills2019stochastic,deroos2019active,wenger2020problinsolve}. In Sec.~\ref{sub:probabilistic_linear_algebra} we will connect this to \gp~inference for a special kernel. Inference in such models has cost $\mathcal{O}(N^2D + N^3)$.

Extending classic quasi-Newton algorithms to a nonparameteric Hessian estimate has been done by \citet{hennig2013quasi} and followed up by \citet{hennig2013noisygradients}. The authors modeled the elements of the Hessian using a high-dimensional \gp~with the RBF kernel and a special matrix-variate structure to allow cost-efficient inference. They also generalized the traditional secant equation to integrate the Hessian along a path for observations, which was possible due to the closed-form integral expression of the RBF kernel. \citet{wills2017construction} expanded this line of work in two directions. They used the same setup as \citet{hennig2013quasi} but explicitly encoded symmetry of the Hessian estimate. The authors also considered modeling the joint distribution of function, gradient and Hessian ($[f,\v{g},H]$) for system identification in the presence of significant noise, and where the computational requirement of inference was less critical. In Section \ref{sub:optimization} we present two similar optimization strategies that utilize exact efficient gradient inference for nonparametric optimization.

\section{Applications}
\label{sec:applications}
Section~\ref{sec:theory} showed how gradient inference for \gp s can be considerably accelerated when $N<D$.
We outline three applications that rely on gradients in high dimensions and that can benefit from a gradient surrogate: optimization, probabilistic linear algebra, and sampling.

\subsection{Optimization} 
\label{sub:optimization}
Unconstrained optimization of a scalar function $f(\v{x}):\R^D \to \R$ consists of locating an input $\v{x}_*$ such that $f(\v{x}_*)$ attains an optimal value, here this will constitute a minimum. This occurs at a point where $\nabla f(\v{x}_*)=\v{0}$.
We focus on Hessian inference from gradients in quasi-Newton methods and then suggest a new method that allows inferring the minimum from gradient evaluations. Pseudocode for an optimization algorithm that uses the inference procedure is available in Alg.~\ref{alg:optimization}.

\subsubsection{Hessian Inference} 
\label{sub:hessian_inference}
Quasi-Newton methods are a popular group of algorithms that includes the widely known BFGS rule \cite{broyden1970bfgs,fletcher1970bfgs,goldfarb1970bfgs,shanno1970bfgs}.
These algorithms either estimate the Hessian $H(\v{x})=\nabla \nabla\trans f(\v{x})$ or its inverse from gradients.
A step direction at iteration $t$ is then determined as $\v{d}_t=-[H(\v{x}_t)]^{-1}\nabla f(\v{x}_t)$.
\citet{hennig2013quasi} showed how popular quasi-Newton methods can be interpreted as inference with a matrix-variate Gaussian distribution conditioned on gradient information.
Here we extend this idea to the nonparametric setting by inferring the Hessian from observed gradients.
In terms of Eq.~\eqref{eq:lin_gp}, we consider the linear operator $\mathcal{L}$ as the second derivative, i.e.,~the Hessian for multivariate functions.
Once a solution $Z_b^l$ to $(\nabla K \nabla\trans)\vect{Z}=\vect{G}$ has been obtained as presented in Section~\ref{sub:implementation}, it is possible to infer the mean of the Hessian at a point $\v{x}_a$
\begin{equation}
	[\bar{H}(\v{x}_a)]^{ij}=\sum\limits_{bl}(\partial_{a}^i\partial_{a}^j \partial_b^l k) Z_b^l.
	\label{eq:hessian_inference}
\end{equation}
This requires the third derivative of the kernel matrix and an additional partial derivative of Eq.~\eqref{eq:kernel_general} which results in
\begin{equation}
	\begin{split}
		\partial_a^i(\partial_a^j\partial_b^l k(r)) &= k''_{ab} \cdot \Lambda^{jl} (\partial_{a}^{i}r) +k''_{ab} \cdot \Lambda^{il} (\partial_{a}^{j}r) \\
		&+ k''_{ab} (\partial_{a}^{i}\partial_{a}^{j}r)(\partial_{b}^{l}r)\\
		&+ k_{ab}'''(\partial_{a}^{i}r)(\partial_{a}^{j}r)(\partial_{b}^{l}r).
	\end{split} 
\end{equation}

With these derivatives, the posterior mean of the Hessian in Eq.~\eqref{eq:hessian_inference} takes the form
\begin{equation}
\bar{H}(\v{x}_a)=
  \begin{bmatrix}
    \Lambda\tilde{X},\Lambda Z
  \end{bmatrix}
  \begin{bmatrix}
    M & \hat{M}\\
    \hat{M} & 0
  \end{bmatrix}
  \begin{bmatrix}
    \tilde{X}\trans\Lambda\\ 
    Z\trans \Lambda
  \end{bmatrix}
  + \Lambda \cdot \op{Tr}(\breve{M})
  \label{eq:posterior_hessian}
\end{equation}
with $M$, $\hat{M}$ and $\breve{M}$ diagonal matrices of size $N\times N$ containing expressions of $k''$ and $k'''$, found in Appendix~\ref{A:gh_inference} alongside a derivation.
$\tilde{X}$ is either $(\v{x}_a - X)$ for stationary kernels or $(X-\v{c})$ for the dot-product kernels.
The posterior mean of the Hessian is of diagonal + low-rank structure for a diagonal $\Lambda$, which is common for quasi-Newton algorithms. The matrix inversion lemma, Eq.~\eqref{eq:woodbury}, can then be applied to efficiently determine the new step direction. On a high level this means that once $Z$ in Eq.~\eqref{eq:solution} has been found, then the cost of inferring the Hessian with a \gp~and inverting it is similar to that of standard quasi-Newton algorithms. 

\subsubsection{Inferring the Optimum} 
\label{sub:inferring_optimum}
The standard operation of Gaussian process regression is to learn a mapping $f(\v{x}):\R^D\to \R$. With the gradient inference it is now possible to learn a nonparametric mapping $\v{g}(\v{x}):\R^{D}\to \R^D$, \emph{but} this mapping can also be reversed to learn an input that corresponds to a gradient. In this way it is possible to learn $\v{x}(\v{g})$ and we can evaluate where the model believes $\v{x}(\v{g}=0)$, i.e, the optimum $\v{x}_*$, lies to construct a new step direction. The posterior mean of $\v{x}_*$ conditioned on the evaluation points $X$ at previous gradients $G$ is
\begin{equation}
\begin{split}
	\bar{\v{x}}_* 
	&= \v{x}_t + [\nabla K(0,G)\nabla] (\nabla K(G,G)\nabla)^{-1}(X-\v{x}_t)\\
	&= \v{x}_t + \Lambda ZK_{b*}' +  \Lambda \tilde{G}(K_{b*}''\odot(Z\trans \Lambda \tilde{\v{g}}_*)).
\end{split}
	\label{eq:solution_inference}
\end{equation}
Here we included a prior mean in the inference which corresponds to the location of the current iteration $\v{x}_t$ and all the inference has been flipped, i.e., gradients are inputs to the kernel and previous points of evaluation are observations. This leads to a new step direction determined by $\v{d}_{t+1}=\bar{\v{x}}_* - \v{x}_t$. For dot product kernels $\tilde{G}\in\R^{D\times N}=G - \v{c}$ and $\tilde{\v{g}}_*=-\v{c}$. 
For stationary kernels $\tilde{G}=(\v{g}_* - G)=-G$ and $Z\trans \Lambda \tilde{\v{g}}_*$ is replaced by $\sum_l Z_b^l\cdot (\Lambda G)_b^l$, derivations in Appendix~\ref{A:infer_min}.

\subsection{Probabilistic Linear Algebra} 
\label{sub:probabilistic_linear_algebra}
Assume the function we want to optimize is
\begin{equation}
	f(\v{x})=\frac{1}{2} (\v{x}-\v{x}_*)\trans A (\v{x}-\v{x}_*),
	\label{eq:Axb}
\end{equation}
with $A\in\R^{D\times D}$ a symmetric an positive definite matrix. Finding the minimum is equivalent to solving the linear system $A\v{x}=\v{b}$, because the gradient $\nabla f(\v{x})=A(\v{x}-\v{x}_*)$ is zero when $A\v{x}=A\v{x}_* \coloneqq \v{b}$.
To model this function we use the second order polynomial kernel
\begin{equation*}
	k(\v{x}_a,\v{x}_b)=\frac{1}{2}\left[(\v{x}_a-\v{c})\trans \Lambda (\v{x}_b-\v{c})\right]^2,
\end{equation*}
and we include a prior mean of the gradient $\v{g}_c=\nabla f(\v{c})=A(\v{c}- \v{x}_*)$. For this setup the overall computational cost decreases from $\mathcal{O}(N^2D + (N^2)^3)$ to $\mathcal{O}(N^2D + N^3)$ because Eq.~\eqref{eq:inner} has the analytical solution
\begin{equation*}
 	Q = \frac{1}{2} (\tilde{X}\trans \Lambda \tilde{X})^{-1} (\tilde{X}\trans A \tilde{X}),
\end{equation*} 
which only requires the inverse of an $N\times N$ matrix instead of an $N^2\times N^2$.
The appearance of $(\tilde{X}\trans A \tilde{X})$ stems from
\begin{equation*}
\begin{split}
	\tilde{X}\trans (G-\v{g}_c)
	&=\tilde{X}\trans (A(X-\v{x}_*)-A(\v{c}-\v{x}_*))=\\
	&=\tilde{X}\trans (A(X -\v{c}))=\tilde{X}\trans A \tilde{X}.
\end{split}
\end{equation*}

It is now possible to apply the Hessian and optimum inference from Sec.~\ref{sub:optimization} specifically to linear algebra at reduced cost. 
If the Hessian inference (cf.~Sec.~\ref{sub:hessian_inference}) is used in this setting, then it leads to a matrix-based probabilistic linear solver \cite{bartels2019unifying,hennig2015probabilistic}.
If instead the reversed inference on the optimum is used (cf.~Sec.~\ref{sub:inferring_optimum}), it will lead to an algorithm reminiscent of the solution-based probabilistic linear solvers \cite{bartels2019unifying,cockayne2019bayesian}.
A full comparison is beyond the scope of this paper and is left for future work.

\begin{algorithm}[tb]
   \caption{GP-[{\textcolor{cH}H}/{\textcolor{cX}X}] Optimization}
   \label{alg:optimization}
\begin{algorithmic}
	\REQUIRE $x_0$
	\STATE {\bfseries Input:} data $\v{x}_0$, $f(\cdot)$,gradient $g(\cdot)$, kernel $k$,size $m$
	\STATE $\v{d}_0=-g(\v{x}_0)$
	\REPEAT
	\STATE $\alpha$ = LineSearch($\v{d}_t$, $f(\cdot)$, $g(\cdot)$)
	\STATE $\v{x}_t$ += $\alpha \v{d}_t$
	\STATE $\v{g}_t$ = g($\v{x}_t$)
	\STATE \textcolor{cH}{$H_t$ = inferH($\v{x}_t\mid X,G$)} \hfill\COMMENT{Eq.~\eqref{eq:posterior_hessian}}
	\STATE \textcolor{cH}{$\v{d}_t$ = -$H_t^{-1}\v{g}_t$} \hfill\COMMENT{quasi-Newton step}
	\STATE updateData($k$, $\v{x}_t$, $\v{g}_t$) \hfill\COMMENT{Keep last $m$ observations}
	\STATE \textcolor{cX}{$\v{d}_t$ = inferMin($\v{0}\mid X-\v{x}_t,G$)} \hfill\COMMENT{Eq.~\eqref{eq:solution_inference}}
	\IF{$\v{d}_t\trans \v{g}_t > 0$}
	\STATE $\v{d}_t$ = $-\v{d}_t$ \hfill \COMMENT{Ensure descent}
	\ENDIF
	\UNTIL{converged}
\end{algorithmic}
\end{algorithm}

\subsection{Hamiltonian Monte Carlo} 
\label{sub:hmc}

Hamiltonian Monte Carlo (\hmc) \citep{duane1987hmc}, is a Markov chain Monte Carlo algorithm that overcomes random walk behavior by introducing gradient information into the sampling procedure \citep{neal2012mcmc, betancourt2018conceptual}.
The key idea is to augment the state space by a momentum variable $\v{p}$ and simulate the dynamics of a fictitious particle of mass $m$ using Hamiltonian mechanics from physics in order to propose new states.
The potential energy $E$ of states $\v{x}$ relates to the target density $P$ as $P(\v{x}) \propto \exp(-E(\v{x}))$.
The Hamiltonian represents the overall energy of the system, i.e., ~potential and kinetic contributions
\begin{equation}
 H(\v{x}, \v{p}) = E(\v{x}) + \frac{\v{p}\Trans \v{p}}{2m},
\end{equation}
and is a conserved quantity.
The joint density of $\v{x}$ and $\v{p}$ is $P(\v{x}, \v{p}) \propto \operatorname{e}^{-H(\v{x}, \v{p}) }$.
Hamiltonian dynamics are solutions to the Hamiltonian equations of motion
\begin{equation}
 \frac{\mathrm{d} \v{x}}{ \mathrm{d}t} = \nabla_{\v{p}} H = \frac{\v{p}}{m}\quad\text{and}\quad \frac{\mathrm{d} \v{p}}{ \mathrm{d}t} = \nabla_{\v{x}} H = - \nabla_{\v{x}} E.
\end{equation}
New states are proposed by numerically simulating trajectories for $T$ steps and stepsize $\epsilon$ using a leapfrog integrator which alternates updates on $\v{p}$ and $\v{x}$.
The theoretical acceptance rate is 1 due to energy conservation; in practice, the discrete solver alters the energy and the new state is accepted or rejected according to a standard Metropolis acceptance criterion.
In this iterative procedure, the gradient of the potential energy $\nabla_{\v{x}} E$ (but not $E$ itself) has to be evaluated repeatedly.\\
\hmc~is therefore inappropriate to simulate from probabilistic models in which the likelihood is costly to evaluate.
This setting arises, e.g.,~when evaluations of the likelihood rely on simulations, or for large datasets $y_{1:M}$ such that $E(\v{x})\! = -\!\sum_{i=1}^M \log p(y_i \mid \v{x}) - \log p(\v{x})$.
For the latter case, subsampling has been proposed. It gives rise to stochastic gradients, but still produces valid states \citep{welling2011bayesian, chen2014stochastic}.
\citet{betancourt2015} advises against the use of subsampling in \hmc~because the discrepancy between the true and subsampled gradient grows with dimension. 
Wrong gradients may yield trajectories that differ significantly from trajectories of constant energy and yield very low acceptance rates and thus, poor performance.

An alternative is to construct a surrogate over the gradient of the potential energy. \citet{rasmussen2003hybridmc} used \gp s to jointly model the potential energy and its gradients. More recently, \citet{Li2019NNG-HMC} obtained better performance with a shallow neural network that is trained on gradient observations during early phases of the sampling procedure.
With novel gradient inference routines we revisit the idea to replace $\nabla_{\v{x}} E$ by a \gp~gradient model that is trained on spatially diverse evaluations of the gradient during early phases of the sampling.

\section{Experiments} 
\label{sec:experiments}
In the preceding section we outlined three applications where nonparametric models could benefit from efficient gradient inference in high dimensions.
These ideas have been explored in previous work with the focus of improving traditional baselines, but always with various tricks to circumvent the expensive gradient inference.
Since the purpose of this paper is to enable gradient inference and not develop new competing algorithms, the presented experiments are meant as a proof-of-concept to assess the feasibility of high-dimensional gradient inference for these algorithms. To this end, the algorithms only used available gradient information in concordance with the baseline.
Details and parameters for reproducibility of all experiments are in Appendix~\ref{A:experiments}.

\subsection{Linear Algebra} 
\label{sub:linear_algebra}

\begin{figure}[t]
\centering
\includegraphics[width=\columnwidth]{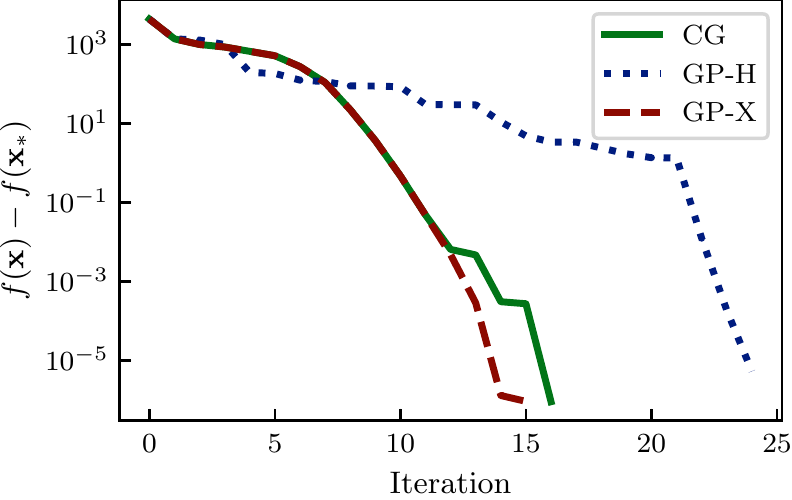}
\caption{Optimization of a 100-dimensional quadratic function, Eq.~\eqref{eq:Axb}, using Alg.~\ref{alg:optimization} with a quadratic kernel as outlined in Sec.~\ref{sub:probabilistic_linear_algebra}. The new solution-based inference shows performance similar to \cg. The presented Hessian-based algorithm uses a fixed $\v{c}=0$ which compromises the performance.}
\label{fig:linear}
\end{figure}

Consider the linear algebra, i.e.,~quadratic optimization, problem in Eq.~\eqref{eq:Axb}.
Quadratic problems are ubiqitous in machine learning and engineering applications, since they form a cornerstone of nonlinear optimization metods.
In our setting, they are particularly interesting due to the computational benefits highlighted in section \ref{sub:probabilistic_linear_algebra}.
There has already been plenty of work studying the performance of probabilistic linear algebra routines \cite{wenger2020problinsolve,bartels2019unifying,cockayne2019bayesian}, of which the proposed Hessian inference for linear algebra is already known \cite{hennig2015probabilistic}. We include a synthetic example of the kind Eq.~\eqref{eq:Axb} to test the new reversed inference on the solution in Eq.~\eqref{eq:solution_inference}. 
Figure \ref{fig:linear} compares the convergence of the gold-standard method of conjugate gradients (\cg) \cite{hestenes1952methods} with Alg.~\ref{alg:optimization} using the efficient inference of section \ref{sub:probabilistic_linear_algebra}.
The matrix $A$ was generated to have spectrum with approximately the $30$ largest eigenvalues in $[1,100]$ and the rest distributed around $0.5$.
The \gp-algorithm retained all the observations to operate similarly to other probabilistic linear algebra routines.
In particular, the probabilistic methods also the optimal step length $\alpha_i = -\v{d}_i\trans \v{g}_i/\v{d}_i\trans A \v{d}_i$ that is used by \cg.

\subsection{Nonlinear Optimization} 
\label{sub:nonlinear_optimization}
\begin{figure}[t]
\centering
\includegraphics[width=\columnwidth]{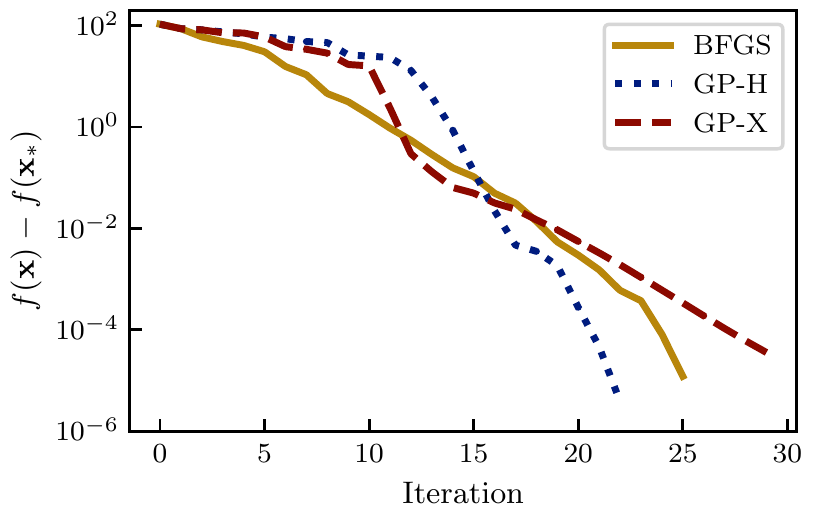}
\caption{Comparison of Alg.~\ref{alg:optimization} with an isotropic RBF kernel against \texttt{scipy}'s implementation of BFGS on a 100-dimensional version of Eq.~\eqref{eq:nonlinear}. All algorithms shared the same line search routine and show similar performance.}
\label{fig:nonlinear}
\end{figure}
The prospect of utilizing a nonparametric model for optimization is more interesting to evaluate in the nonlinear setting. In Fig.~\ref{fig:nonlinear} the convergence of both versions of Alg.~\ref{alg:optimization} are compared to \texttt{scipy}'s BFGS implementation. The nonparametric models use the RBF kernel with the last 2 observations for inference. All algorithms share the same line search routine. The function to be minimized is a relaxed version of a 100-dimensional Rosenbrock function \cite{rosenbrock1960function}

\begin{equation}
	f(\v{x})=\sum_{i=1}^{D-1} x_i^2 + 2\cdot(x_{i+1}-x_i^2)^2.
	\label{eq:nonlinear}
\end{equation}

\begin{figure}[t]
\centering
\includegraphics[width=\columnwidth]{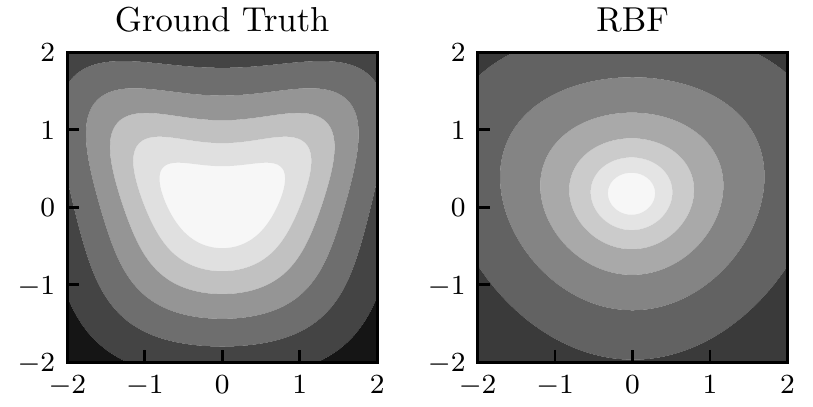}
\caption{The first two dimensions of Eq.~\eqref{eq:nonlinear} along with the inferred curvature from 1000 randomly distributed samples. The inferred function has identified the minimum and a slight elongation of the function but not the minute details of the shape.}
\label{fig:nl_estimate}
\end{figure}

A hyperplane of the function can be seen on the left in Fig.~\ref{fig:nl_estimate} for the first two dimensions with every other dimension evaluated at 0.
The right plot shows the same plane with the function values inferred from gradient observations evaluated at $N=1000$ uniformly randomly distributed evaluations in the hypercube $\v{x}_n^i\in[-2,2]^{100}$.
Constructing the Gram matrix for these observations would require $(1000\cdot100)^2$ floating point numbers, which for double precision would amount to $>74\,$GB of memory.
Instead the multiplication in Eq.~\eqref{eq:matvec} was used in conjunction with an iterative linear solver to approximately solve the linear system. This approach required storage of $3ND + 3 N^2$ numbers (\cg~requirements and intermediate matrices included) amounting to a total of only 25\,MB of RAM.
The solver ran for 520 iterations until a relative tolerance of $10^{-6}$ was reached, which took ~4.9 seconds on a 2.2GHz 8-core processor. Extrapolating this time to $100\cdot 1000$ iterations (the time to theoretically solve the linear system exactly) would yield approximately 16 minutes. Such iterative methods are sensitive to roundoff errors and are not guaranteed to converge for such large matrices without preconditioning. The required number of iterations to reach convergence vary with the lengthscale of the kernel and chosen tolerance. For this experiment a lengthscale of $\ell^2=10\cdot D$ was used with the isotropic RBF kernel, i.e., the inverse lengthscale matrix $\Lambda=10^{-3}\cdot I$.

\subsection{Gradient Surrogate Hamiltonian Monte Carlo} 
\label{sub:hmc_exp}

\begin{figure}[t]
\centering
\includegraphics[width=\columnwidth]{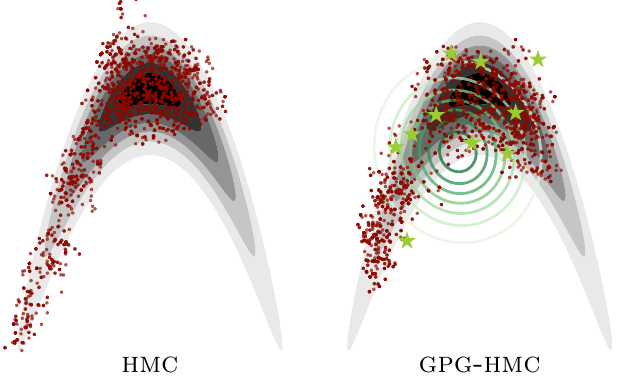}
\caption{2000 samples drawn with \hmc~\emph{(left)} and \gpghmc~on a 100 dimensional problem. Displayed is a projection onto 2 dimensions, all other dimensions are Gaussian. 
Acceptance rates are 0.51 (\hmc) and 0.39 (\gpghmc). \gpghmc~uses 372 iterations with \hmc~with acceptance rate 0.57 of which it selects 10 points for training, displayed as {\protect\tikz \protect\node[star,star points=5, star point ratio=2.5, fill=yellowgreen, scale=0.5] {};}. Elliptical contours indicate the posterior on the target inferred from the 10 gradient evaluations.}
\label{fig:banana}
\end{figure}

For \hmc, we take a similar approach to the one taken by \citet{rasmussen2003hybridmc} and build a global surrogate model on the gradient of the potential energy $E(\v{x}) = -\log P(\v{x})$.
Our model differs in that we predict the gradient $\nabla E$ only from previous gradient observations, not including function evaluations as does \citet{rasmussen2003hybridmc}.

We construct a synthetic 100 dimensional target density that is banana-shaped in two dimensions and Gaussian in all the other dimensions.
Fig.~\ref{fig:banana} shows the conditional density in the two non-Gaussian dimensions, together with projected samples that were collected using standard \hmc~and \hmc~using a \gp~gradient surrogate, which we denote as \gpghmc.
For training of \gpghmc, we assign a budget $N = \lfloor\sqrt{D}\rfloor$ and run \hmc~until $N/2$ points are found that are more than a kernel lengthscale apart.
Then we switch into the surrogate mode in which the true $\nabla E$ is queried only if a new location sufficiently far from the previous ones is found to condition the \gp~on until the budget is reached.

In Fig.~\ref{fig:banana}, the isotropic square exponential kernel is aligned with the intrinsic dimensions of the problem.
Therefore, we also consider 10 arbitrary rotations of the same problem by applying a random orthonormal matrix on the input and repeat each configuration for 10 different initializations.
We find that over 2000 samples, \hmc~has an acceptance rate of $0.46 \pm 0.02$ and \gpghmc~achieves $0.50 \pm 0.02$ using the gradient surrogate. \gpghmc~was conditioned on $N=10$ gradient observations collected during the first $650 \pm 82$ iterations of \hmc. The higher acceptance rate is related to the mismatch between the estimated and true gradient that tends to cause a skewed distribution of $\Delta H$ towards positive values.
The acceptance criterion still queries the true potential energy $E$, thus \gpghmc~produces valid samples of $\operatorname{e}^{-E}$.
As $E$ gets increasingly expensive to evaluate, \gpghmc~thus offers a lightweight surrogate that drastically reduces the number of calls to the true gradient.

\section{Discussion and Future Work} 
\label{sec:discussion}

We have presented how structure inherent in the kernel Gram matrix can be exploited to lower the cost of \gp~inference with gradients from cubic to linear in the dimension.
This technical observation principally opens up entirely new perspectives for high-dimensional applications in which gradient inference has previously been dismissed as prohibitive.
We demonstrate on a conceptual level the great potential of this reformulation on various algorithms.
The major intention behind the paper, however, is to spark research to overhaul algorithms that operate on high-dimensional spaces and leverage gradient information.

The speed-up in terms of dimensionality does not come without limitations.
Our proposed decomposition compromises the number of permissible gradient evaluations compared to the na\"ive approach to gradient inference.
Hence, our method is applicable only in the low-data regime in which $N < D$.
This property is unproblematic in applications that benefit from a local gradient model, e.g.,~in optimization.
Nevertheless, we also found a remedy for the computational burden when $N > D$ using iterative schemes.
Furthermore, the structure we uncovered allows storing the quantities that are necessary to multiply the Gram matrix with an arbitrary vector.
We thus showed that global models of the gradient are possible when a low-confidence gradient belief is sufficient.
This is of particular interest for \gp~implementations that leverage the massive parallelization available on GPUs where available memory often becomes the bottleneck.

The most efficient numerical algorithms use knowledge about their input to speed up the execution.
Explicit structural knowledge is usually reflected in hard-coded algorithms, e.g., linear solvers for matrices with specific properties that are known a priori.
Structure can also be included in probabilistic numerical methods where the chosen model encodes known symmetries and constraints. 
At the same time, these methods are robust towards numeric uncertainty or noise, which can be included in the probabilistic model.
Since Gaussian processes form a cornerstone of probabilistic numerical methods \citep{hennig2015pn}, our framework allows the incorporation of additional functional constraints into numerical algorithms for high-dimensional data.
Actions taken by such algorithms are then better suited to the problem at hand.
The cheap inclusion of \gp~gradient information in numerical routines might therefore enable new perspectives for algorithms with an underlying probabilistic model.

\section*{Acknowledgements}
AG acknowledges funding by the European Research Council through ERC StG Action 757275 / PANAMA. Both FdR and AG thank the International Max Planck Research School for Intelligent Systems (IMPRS-IS) for support.

\bibliography{bibliography}
\bibliographystyle{icml2021}

\clearpage
\appendix

\onecolumn
\icmltitle{Supplementary Material\\ \large{High-Dimensional Gaussian Process Inference with Derivatives}}
\section{Linear Algebra}
\label{A:la}
Kronecker products play an important role in the derivations so here list a few properties that will be useful, see \cite{vanloan2000kronecker} for more. The Kronecker product for a matrix $A\in \R^{M\times N}$ and $B\in \R^{P\times Q}$ is a block matrix $(A\otimes B)\in \R^{MP\times NQ}$ with block $[i,j]=A_{ij}\cdot B$. We will also require the ``perfect shuffle'' matrix $S$ and the column-stacking operation of a matrix  $\vect{\cdot}$ \cite{vanloan2000kronecker}.

\paragraph*{Properties}
For matrices of appropriate sizes (these will be valid for the derivations).
\begin{itemize}
  \item $(A\otimes B)^{-1}=(A^{-1}\otimes B^{-1})$
  \item $(A\otimes B)(C\otimes D)=(AC\otimes BD)$
  \item $S_{NQ}\vect{X}=\vect{X\trans}$ for $X\in\R^{Q\times N}$
  \item $(A\otimes B)\vect{X}=\vect{BXA\trans}$ for $A\in \R^{M\times N}$, $B\in \R^{P\times Q}$ and $X\in\R^{Q\times N}$.
\end{itemize}
The final property is particularly prevalent the in derivations so we introduce the shorthand 
$$(A\otimes B)\vect{X}\rightarrow BXA\trans$$
to denote the "unvectorized" result. If the vectorization operation is applied to the result then the flattened correct result is obtained.

\paragraph*{Notation}
The derivations contain several several matrices that we here list to give an overview. The input dimension is $D$ and there are $N$ observations.
\begin{itemize}
  \item $X\in\R^{D\times N}$: All evaluation points stacked into a matrix.
  \item $\nabla K \nabla'\in\R^{DN\times DN}$: Kernel gram matrix for the derivatives with decompositions $\nabla K \nabla'=B + UCU\trans$.
  \item $G\in \R^{D\times N}$: All gradients stacked into a matrix. $\vect{G}$ r.h.s. of $\nabla K \nabla' \vect{Z}=\vect{G}$. 
  \item $Z\in \R^{D\times N}$: the solution to $\nabla K \nabla' \vect{Z}=\vect{G}$, (Riesz representers).
  \item $B\in \R^{DN\times DN}$: Kronecker product of $K' \otimes \Lambda$ 
  \item $C\in \R^{N^2\times N^2}$: Symmetric matrix defined as $C=\op{diag}(\vect{K''})S_{NN}=S_{NN}\op{diag}(\vect{K''})$.
  \begin{itemize}
    \item $C\vect{M}\rightarrow K''\odot M\trans$.
    \item $C^{-1}\vect{M}\rightarrow M\trans \oslash K''$. 
    \item $\odot$ and $\oslash$ correspond to the elementwise multiplication and division respectively.
  \end{itemize} 
  \item $U\in \R^{ND\times N^2}$: Tall and thin Kronecker product used in $\nabla K \nabla'=B+UCU\trans$.
  \begin{itemize}
    \item For dot product kernels $U=(I\otimes \Lambda(X-\v{c}))$.
    \item For stationary kernels $U=(I\otimes\Lambda X)L$.
  \end{itemize}
  \item $L\in \R^{N^2\times N^2}$: Sparse operator required for $U$ in stationary kernels 
  \begin{itemize}
    \item $[L\trans \vect{M}]_{ab}\rightarrow M_{aa}-M_{ab}$ 
    \item $[L\vect{M}]_{ab}\rightarrow \op{diag}(\sum\limits_{a}M_{ab})-M_{ab}$
  \end{itemize}
  \item $Q\in \R^{N\times N}$: Solution to $(C^{-1} + U\trans B^{-1}U)\vect{Q}=U\trans B^{-1} \vect{G} $
  
  \end{itemize}

\section{Kernel Derivatives}
\label{A:kernel_derivatives}
Conditioning a GP on gradient observations requires the derivative of the kernel w.r.t. its arguments.
Here we derive these terms for kernels with inner products and stationary kernels. We use the notation ${\partial_{b}}^j$ as shorthand for $\partial/\partial {\v{x}_b}^j$ and use $k'$ to refer to the derivative w.r.t. the scalar argument $r$.
The notation mirrors that of Sec.~\ref{sec:theory}.

\subsection{General Kernels}
\label{A:general}
If we write a general kernel $k(\v{x}_a,\v{x}_b)=k(r(\v{x}_a,\v{x}_b))$ then the general form of each component for gradient inference will take the following form.

\begin{equation}
\begin{split}
k(\v{x}_a,\v{x}_b) &= k(r(\v{x}_a,\v{x}_b))\\
{\partial_{b}}^{j}k(r) 
&= k'(r) {\partial_{b}}^{j} r\\ 
{\partial_{a}}^{i}k(r) 
&= k'(r) {\partial_{a}}^{i} r\\
{\partial_a}^i {\partial_b}^j k(r) 
&= k_{ab}'(r)\cdot {\partial_a}^i {\partial_b}^j r  + k_{ab}''(r)\cdot ({\partial_a}^i r) ({\partial_b}^j r)
\end{split}
\label{eq:Akernel_derivative}
\end{equation}

We thus use the convention of ordering the entries in the Gram matrix $\nabla K \nabla'$ first according to the $N$ data points $\v{x}_{1:N}$, and then according to dimension, i.e.,
\begin{equation}
\nabla K \nabla' = 
 \begin{pmatrix}
  \nabla k(\v{x}_1, \v{x}_1) \nabla' & \dots & \nabla k(\v{x}_1, \v{x}_N) \nabla'\\
  \vdots & \ddots & \vdots\\
  \nabla k(\v{x}_N, \v{x}_1) \nabla' & \dots & \nabla k(\v{x}_N, \v{x}_N) \nabla'\\
 \end{pmatrix},
\end{equation}

where each block has the size $D\times D$.
We highlight this ordering as it deviates from the conventional way found in the literature.
Each element of the $a,b^{\text{th}}$ block take the form ${\partial_a}^i {\partial_b}^j k(r)$ specified in Eq.~\eqref{eq:Akernel_derivative}, where no assumption on the structure of the kernel has been done at this point.
The first term decomposes into a Kronecker product for the kernels we consider, because indices $a,b$ and $i,j$ separate. 
This term can thus be efficiently inverted.
The second term is what usually makes closed-form gradient inference intractable which will be further explored below for dot product kernels and stationary kernels.

\subsection{Dot Product Kernels} 
\label{A:dot_product}
For dot product kernels we define the function $r$ as
\begin{equation}
	r(\v{x}_a,\v{x}_b)=(\v{x}_a- \v{c} )\trans\Lambda(\v{x}_b - \v{c} ).
  \label{eq:Adotprod_def}
\end{equation}
See Sec.~\ref{A:dotprod_examples} for examples of dot product kernels.

The relevant terms of Eq.~\eqref{eq:Akernel_derivative} are:
\begin{equation*}
	\begin{split}
		{\partial_a}^{i} r(\v{x}_a,\v{x}_b) &= [\Lambda(\v{x}_b - \v{c})]^{i}\\
    {\partial_b}^{j} r(\v{x}_a,\v{x}_b) &= [\Lambda(\v{x}_a - \v{c})]^{j}\\
    {\partial_a}^{i}\partial_b^{j} r(\v{x}_a,\v{x}_b) &= \Lambda^{ij}
	\end{split}
\end{equation*}

From this we see the Gram matrix of Eq.~\eqref{eq:Akernel_derivative} will look like:
\begin{equation}
\begin{split}
	{\partial_a}^i {\partial_b}^j k(r) &= k_{ab}'(r)\cdot\Lambda^{ij} + k_{ab}''(r)\cdot [\Lambda(\v{x}_b - \v{c} )]^{i} [(\v{x}_a - \v{c} )\trans \Lambda]^{j} \\
	&= [K \otimes \Lambda]_{ab}^{ij} + [(I\otimes \Lambda\tilde{X})\underbrace{
  (S_{NN}\op{diag}(\vect{K''}))
  }_{C}(I\otimes \tilde{X}\Lambda)\trans ]_{ab}^{ij}
\end{split}
\end{equation}
The first term is of Kronecker structure which is easy to invert using properties of Kronecker products.
The second consists of rank-1 corrections block-wise multiplied with the scalar value $k_{ab}''$. The input indices are flipped for the term i.e., $b$ appears as a row index and $a$ as column. This shuffling is what makes the structure of the gradient Gram matrix difficult, but it can be resolved with the Kronecker transposed product.
To derive the structure of the second term we start by defining the matrix $\tilde{X}\in\R^{D\times N}$,  $\tilde{X}=X-\v{c}$. 
We can then form the following outer product to get the structure:

\begin{equation*}
  \begin{split}
    \left[\Lambda(\v{x}_b - \v{c}) \right]^i \left[(\v{x}_a - \v{c})\trans \Lambda\trans\right]^j &= [\Lambda \tilde{X}_b]^i [(\Lambda \tilde{X}_b)\trans]^j\\
    &= \sum_{m,n}^N [\Lambda\tilde{X}_{n}]^i [\Lambda \tilde{X}_{m}]^j \delta_{am}\delta_{bn}\\
    &= \sum_{n, n'}^N \sum_{m, m'}^N [\Lambda\tilde{X}_{n}]^i [\Lambda \tilde{X}_{m}]^j \delta_{am'}\delta_{bn'} \delta_{mm'}\delta_{nn'}\\
    &= \sum_{n, n'}^N \sum_{m, m'}^N \left(\delta_{am'} \cdot [\Lambda\tilde{X}_{n}]^i\right) 
    \underbrace{\left(\delta_{mm'}\delta_{nn'}\right)}_{S_{NN}} 
    \left(\delta_{bn'} \cdot [\Lambda \tilde{X}_{m}]^j \right)\\
    &= \sum_{n, n'}^N \sum_{m, m'}^N  [I \otimes \Lambda\tilde{X}]^{i}_{a,m'n}\, [
    S_{NN}
    ]_{m'n,n'm}\, [I \otimes (\Lambda\tilde{X})\trans]^{j}_{n'm,b}\\
    &= \left[(I\otimes \Lambda\tilde{X}) 
    S_{NN}
    (I\otimes \tilde{X}\Lambda)\trans \right]^{ij}_{ab}\\
	\end{split}
\end{equation*}

To get the right scalar value for each block outer product one has to write the term like below.
\begin{equation}
\underbrace{(I\otimes \Lambda \tilde{X})}_{U} 
\underbrace{(S_{NN}\op{diag}(\vect{K''}))}_{C} 
\underbrace{(I\otimes \Lambda \tilde{X})\trans}_{U\trans}
\label{eq:AdotUCU}
\end{equation} 
with $C_{m'n,n'm}=K_{mn}'' \delta_{mm'}\delta_{nn'}$ a symmetric $N^2 \times N^2$ matrix.

\subsubsection{Examples for Inner Product Kernels}
\label{A:dotprod_examples}

\begin{table}[h!]
  \centering
  {\renewcommand{\arraystretch}{1.5}
  \begin{tabular}{lccc}
  Kernel &  $k(r)$ &  $k'(r)$ &  $k''(r)$ \\
  \hline\hline
  Polynomial($p$) & $\frac{r^p}{p(p-1)}$ & $\frac{r^{p-1}}{(p-1)}$ & $r^{p-2}$  \\
  Polynomial(2) & $\frac{r^2}{2}$ & $r$ & $1$  \\
  Exponential/Taylor & $\exp\left(r \right)$ & $\exp\left(r \right)$ & $\exp\left(r\right)$  \\
  \end{tabular}
  }
  \caption{Examples for inner product kernels where $r=(\v{x}_a- \v{c} )\trans\Lambda(\v{x}_b - \v{c} )$.}
\end{table}

\subsection{Stationary Kernels}
\label{A:stationary}

For a stationary kernel we define
\begin{equation*}
	r(\v{x}_a,\v{x}_b)=(\v{x}_a- \v{x}_b )\trans\Lambda(\v{x}_a - \v{x}_b ).
\end{equation*}
Note here the discrepancy to conventional notation and do \emph{not} think of $r$ as a radius or Mahalonobis distance here (but rather its square).
Then we have the following identities:
\begin{equation*}
	\begin{split}
		{\partial_a}^i r(\v{x}_a,\v{x}_b) 
    &= 2\cdot [\Lambda(\v{x}_a - \v{x}_b )]^i\\
		{\partial_b}^j r(\v{x}_a,\v{x}_b)  
    &= -2\cdot [\Lambda(\v{x}_a - \v{x}_b )]^j \\
		{\partial_a}^i {\partial_b}^j r^(\v{x}_a,\v{x}_b) 
    &= -4\cdot \Lambda^{ij}.
	\end{split}
\end{equation*}
The Gram matrix will have the general structure:
\begin{equation}
	{\partial_a}^i {\partial_b}^j k(r) = -2k_{ab}'(r)\cdot\Lambda_{jl} - 4k_{ab}''(r) \cdot [\Lambda(\v{x}_a - \v{x}_b )]^{i} [(\v{x}_a - \v{x}_b )\trans \Lambda]^{j}. 
\end{equation}
Usually the factors 2 and 4 disappear due to scalar values of $k'(r)$ and $k''(r)$, see Sec.~\ref{A:stationary_examples}.

Writing the second term in matrix form is a bit more intricate than Eq.~\eqref{eq:AdotUCU}, but taking the same approach we get
\begin{equation}
 \begin{split}
  &[\Lambda(\v{x}_a - \v{x}_b )]^{i} [(\v{x}_a - \v{x}_b )\trans \Lambda]^{j}  
   = [\Lambda\v{x}_a]^i [\v{x}_a\trans\Lambda]^j - [\Lambda\v{x}_b ]^{i} [\v{x}_a\trans\Lambda]^j - [\Lambda\v{x}_a]^i [\v{x}_b\trans \Lambda]^{j} + [\Lambda\v{x}_b ]^{i} [\v{x}_b\trans\Lambda]^j\\
  &= \sum_{mn}  \delta_{am}\delta_{bn} \left( [\Lambda\v{x}_m]^i [\v{x}_m\trans\Lambda]^j - [\Lambda\v{x}_n ]^{i} [\v{x}_m\trans\Lambda]^j -  [\Lambda\v{x}_m]^i [\v{x}_n\trans \Lambda]^{j} + [\Lambda\v{x}_n ]^{i} [\v{x}_n\trans\Lambda]^j \right)\\
  &= \sum_{mn}  \left(\delta_{am} \left( [\Lambda\v{x}_m]^i - [\v{x}_n\trans\Lambda]^i \right)\right) \left(  \delta_{bn} \left([\v{x}_m\trans\Lambda]^j - [\v{x}_n\trans \Lambda]^{j}\right) \right)\\
  &= \sum_{mnpp'}  \left(\delta_{am} \left( \delta_{pm}[\Lambda\v{x}_p]^i - \delta_{pn}[\v{x}_p\trans\Lambda]^i \right)\right) \left(  \delta_{bn} \left(\delta_{p'm}[\v{x}_{p'}\trans\Lambda]^j - \delta_{p'n}[\v{x}_{p'}\trans \Lambda]^{j}\right) \right)\\
  &= \sum_{mnoo'pp'}  \left([\Lambda\v{x}_p]^i \delta_{ao} \delta_{mo} (\delta_{pm} - \delta_{pn})\right) \left( [\v{x}_n\trans\Lambda]^j  \delta_{bo'} \delta_{no'}  (\delta_{p'm} -  \delta_{p'n}) \right)\\
  &= \sum_{mn} \sum_{op}  \underbrace{\delta_{ao} [\Lambda\v{x}_p]^i}_{U_{ai,op}}  \underbrace{\delta_{om} (\delta_{pm} - \delta_{pn})}_{L_{op,mn}}  \sum_{o'p'} \underbrace{\delta_{o'n}  (\delta_{p'm} - \delta_{p'n})}_{L_{mn, o'p'}}\ \underbrace{\delta_{o'b} [\v{x}_n\trans\Lambda]^j}_{U_{o'p', bj}} \\
 \end{split}
\end{equation}
For dot product kernels we used $U=(I\otimes \Lambda(X-\v{c}))$, for stationary kernels we instead use $U=(I\otimes \Lambda X)L$.
The second term of the Gram matrix is formed by $UCU\trans$ in the same way as Eq.~\eqref{eq:AdotUCU}. $U$ is however no longer a Kronecker product which makes the algorithmic details more involved. It is therefore more convenient to use the $UL$ representation where $L$ is a sparse linear operator. $U\trans\vect{g}=L\trans\vect{X\trans \Lambda g}_{mn}= \vect{X\trans \Lambda g_{mn} - X\trans \Lambda g_{mm}}$

\subsubsection{Examples for Stationary Kernels}
\label{A:stationary_examples}

\begin{table}[h!]
  \centering
  {\renewcommand{\arraystretch}{1.5}
  \begin{tabular}{lccc}
  Kernel &  $k(r)$ &  $k'(r)$ &  $k''(r)$ \\
  \hline\hline
  Squared exponential& $\op{e}^{-r/2}$ & $-\frac{1}{2} k(r)$ & $\frac{1}{4} k(r)$  \\
  Mat\'ern $\nu = \nicefrac{1}{2}$ & $\op{e}^{-\sqrt{r}}$  & $ -\frac{k(r)}{2\sqrt{r}}$  & $ \frac{1}{4r^{3/2}} \left(\sqrt{r} + 1 \right) k(r)$   \\
  Mat\'ern $\nu = \nicefrac{3}{2}$ & $(1+\sqrt{3r})\op{e}^{-\sqrt{3r}}$  & $\frac{\sqrt{3}}{2\sqrt{r}} \left(\op{e}^{-\sqrt{3r}} - k(r) \right)$ &   
  $ \frac{\sqrt{3}}{2\sqrt{r}} \left(\frac{k(r)}{2r} - k'(r) - \op{e}^{-\sqrt{3r}}\frac{1 + \sqrt{3r}}{2r}\right)$\\
  Mat\'ern $\nu = \nicefrac{5}{2}$ & $\left(1+\sqrt{5r} + \frac{5r}{3}\right)\op{e}^{-\sqrt{5r}}$  & $\left(\frac{\sqrt{5}}{2\sqrt{r}} + \frac{5}{3}\right)\op{e}^{-\sqrt{5r}} - \frac{\sqrt{5}}{2\sqrt{r}} k(r)$ &   $ \frac{\sqrt{5}}{2\sqrt{r}} \left(\frac{k(r)}{2r} - k'(r) - \op{e}^{-\sqrt{5r}} \left( \frac{1 + \sqrt{5r}}{2r} + \frac{5}{3} \right) \right)$\\
  Rational quadratic & $\left( 1 + \frac{r}{2\alpha} \right)^{-\alpha}$  & $-\frac{1}{2}\left( 1 + \frac{r}{2\alpha} \right)^{-\alpha-1}$ & $\frac{\alpha + 1}{4\alpha}\left( 1 + \frac{r}{2\alpha} \right)^{-\alpha-2}$  \\
  \end{tabular}
  }
  \caption{Examples for stationary kernels where $r=(\v{x}_a- \v{x}_b )\trans\Lambda(\v{x}_a - \v{x}_b )$.}
  \label{tbl:stationary_kernels}
\end{table}

Table~\ref{tbl:stationary_kernels} contains the kernels we considered.
For reasons of space, we derive the general expressions for the Mat\'ern family with half integer smoothness parameter $\nu = p + \frac{1}{2}$ for $p\in \mathbb{N}$ here, which reads
 $$ k_{p + \nicefrac{1}{2}} (r) = \exp \left(-\sqrt{2\nu r}\right) \frac{\Gamma (p+1)}{\Gamma (2p+1)} \sum_{i=0}^p \frac{(p+i) !}{i! (p-i)!} \left(\sqrt{8\nu r}\right)^{p-i}, $$
 and has the monstrous derivatives
 \begin{align*}
  k_{p + \nicefrac{1}{2}}' (r) &= -\sqrt{\frac{\nu}{2r}} k_{p + \nicefrac{1}{2}} + \exp \left(-\sqrt{2\nu r}\right) \frac{\Gamma (p+1)}{\Gamma (2p+1)} \sum_{i=0}^{p-1} \frac{(p+i) !}{i! (p-i-1)!} \left(\sqrt{8\nu r}\right)^{p-i-1} \sqrt{\frac{2\nu}{r}}\\
  k_{p + \nicefrac{1}{2}}'' (r) &= \left(\sqrt{\frac{\nu}{8r^3}} + \frac{\nu}{2r}\right) k_{p + \nicefrac{1}{2}} 
  - \left(\sqrt{\frac{\nu}{2 r^3}} + \frac{2\nu}{r}\right) \exp \left(-\sqrt{2\nu r}\right) \frac{\Gamma (p+1)}{\Gamma (2p+1)} \sum_{i=0}^{p-1} \frac{(p+i) !}{i! (p-i-1)!} \left(\sqrt{8\nu r}\right)^{p-i-1} \\
  &\quad + \sqrt{\frac{2\nu}{r}} \exp \left(-\sqrt{2\nu r}\right) \frac{\Gamma (p+1)}{\Gamma (2p+1)} \sum_{i=0}^{p-2} \frac{(p+i) !}{i! (p-i-2)!} \left(\sqrt{8\nu r}\right)^{p-i-2}.
 \end{align*}

\section{Decomposition Benefits} 
\label{A:decomposition}
In Appendix \ref{A:kernel_derivatives} we showed that $\nabla K \nabla '$ can be written as $B+UCU\trans$, see Appendix \ref{A:la} for summary. 
In Sec.~\ref{sub:implementation} we discussed some benefits of the decomposition that we here explain more in detail.

\subsection{Woodbury Vector for $N<D$}
\label{A:woodbury_smallN}
The decomposition is particularly interesting when the number of observations $N$ is small. In this setting we can employ the matrix inversion lemma, Eq.~\eqref{eq:woodbury} restated here for convenience

\begin{equation*}
  (B+UCU\trans)^{-1}=B^{-1} - B^{-1}U\left( C^{-1} + U\trans B^{-1}U \right)^{-1}U\trans B^{-1}. 
\end{equation*}
If the size of $C$ is smaller than $B$ and $B^{-1}$ is ``cheap'', then the r.h.s. above is computationally beneficial.
The involved matrices are all comparatively large, but by using the important properties of Kronecker products (Appendix \ref{A:la}) it is possible to significantly lower the requirements. 
Here we outline the required operations for a dot product kernel with $\tilde{X}=X-\v{c}$. The operations for stationary kernels are similar but require the additional application of $L$ for each operation involving $U$.
\begin{enumerate}
  \item $T=U\trans B^{-1}\vect{G}\rightarrow \tilde{X}\trans G (K')^{-1} $. 
  \begin{itemize}
     \item $T\in \R^{N\times N}$
   \end{itemize}
  \item Solve: $\left(C^{-1} + U\trans B^{-1}U\right)\vect{Q}=\vect{T}$: $\left( C^{-1} + (K')^{-1}\otimes (\tilde{X}\trans\Lambda \tilde{X}) \right) \vect{Q}=\vect{T}$. 
  \begin{itemize}
    \item $Q\in \R^{N\times N}$
  \end{itemize}
  \item $\vect{Z}=B^{-1}\vect{G} - B^{-1}U \vect{Q}$: $Z=\Lambda^{-1}G(K')^{-1} - XQ(K')^{-1}$. 
  \begin{itemize}
    \item $Z\in\R^{D\times N}$
  \end{itemize}
\end{enumerate}

\paragraph*{Special Case} 
Step 2 in the above procedure is the source of the $\mathcal{O}((N^2)^3)$ scaling in computations.
For the situation outlined in Sec.~\ref{sub:probabilistic_linear_algebra} it is possible to solve the linear system analytically.
A multiplication with the linear system in step 2 for the second order polynomial kernel is performed as
\begin{equation}
  (C^{-1} + (\tilde{X}\Lambda\tilde{X})^{-1}\otimes(\tilde{X}\trans\Lambda\tilde{X}))\vect{V}\rightarrow V\trans + (\tilde{X}\trans\Lambda\tilde{X})V(\tilde{X}\Lambda\tilde{X})^{-1}.
  \label{eq:ACUBU_mult}
\end{equation}
For the outlined situation in Sec.~\ref{sub:probabilistic_linear_algebra} the r.h.s. $T=(\tilde{X}\trans A\tilde{X})(\tilde{X}\trans\Lambda\tilde{X})^{-1}$.

The solution to the linear system is
\begin{equation*}
  Q=\frac{1}{2}(\tilde{X}\trans\Lambda\tilde{X})^{-1}(\tilde{X}\trans A\tilde{X}).
\end{equation*}
This is easily verified by inserting the value for $Q$ in Eq.~\eqref{eq:ACUBU_mult}
\begin{equation*}
  \begin{split}
  Q\trans + (\tilde{X}\trans\Lambda\tilde{X})Q(\tilde{X}\trans\Lambda\tilde{X})^{-1} 
  &=\frac{1}{2}(\tilde{X}\trans A\tilde{X})(\tilde{X}\trans\Lambda\tilde{X})^{-1} + (\tilde{X}\trans\Lambda\tilde{X})[\frac{1}{2}(\tilde{X}\trans\Lambda\tilde{X})^{-1}(\tilde{X}\trans A\tilde{X})](\tilde{X}\trans\Lambda\tilde{X})^{-1}\\
  &=\frac{1}{2}(\tilde{X}\trans A\tilde{X})(\tilde{X}\trans\Lambda\tilde{X})^{-1} + \frac{1}{2}(\tilde{X}\trans A\tilde{X})(\tilde{X}\trans\Lambda\tilde{X})^{-1} \\
  &=(\tilde{X}\trans A\tilde{X})(\tilde{X}\trans\Lambda\tilde{X})^{-1} = T
  \end{split}
\end{equation*}

\subsection{Benefits for General $N$}
\label{A:woodbury_general}
The derived Kronecker structure of the Gram matrix  $\nabla K \nabla'$ in Eq.~\eqref{eq:kernel_general} highlights an important speedup of multiplication. Multiplying a vectorized matrix $V$ of same shape as $G$ with the Gram matrix is obtained by the following computations
\begin{equation*}
	\nabla K \nabla' \vect{V}= \Lambda V K' + \Lambda X (K''\odot V\trans\Lambda X),
\end{equation*}
A full algorithm for multiplication with the Gram matrix is available in Alg.~\ref{alg:matvec}, with modification for stationary kernels written in red.
The advantage of defining such a routine is that the Gram matrix never needs to be built, which reduces the memory requirement from $\mathcal{O}((DN)^2)$ to $\mathcal{O}(DN + N^2)$.

\begin{algorithm}[tb]
   \caption{$\nabla K \nabla'$-MVM}
   \label{alg:matvec}
\begin{algorithmic}
  \REQUIRE $x_0$
  \STATE {\bfseries Input:} ($V\in \R^{D\times N}$, $K'\in \R^{N\times N}$, $K''\in \R^{N\times N}$, $\tilde{X}\in \R^{D\times N}$)
  \STATE $M$ = $\tilde{X}\trans\Lambda V$
  \STATE \textcolor{cX}{$\v{m}$ = $\op{diag}(M)$} \hfill\COMMENT{Multiplication with $L\trans$}
  \STATE \textcolor{cX}{$M$ = $M-\v{m}\trans$} 
  \STATE $M$ = $K''\odot M\trans$
  \STATE \textcolor{cX}{$\v{m}$ = $\sum_a M_{ab}$} \hfill\COMMENT{Multiplication with $L$}
  \STATE \textcolor{cX}{$M$ = $\v{m}\trans -M$}
  \STATE {\bfseries Return:} $\Lambda V K' + \Lambda \tilde{X} M$
\end{algorithmic}
\end{algorithm}

\section{Gradient and Hessian Inference}
\label{A:gh_inference}
Once $Z\in\R^{D\times N}$ has been obtained from solving $\nabla K \nabla' \vect{Z}=\vect{G}$ it is possible to infer the gradient and Hessian at a new point $\v{x}_a$. Note that $a$ is now an index with a single value and $b$ takes $N$ values, so $K_{a b}=\v{k}_{ab}$ is a row vector. 
Inferring the gradient and Hessian at a point $\v{x}_a$ requires the following contractions
\begin{equation}
  \bar{\v{g}}(\v{x}_a)^i =\sum\limits_{bl} [\partial_a^i \partial_b^l k(r)]_{a b}^{il}Z_b^l,
  \label{eq:Ainf_g}
\end{equation}
and 
\begin{equation}
  \bar{H}(\v{x}_a)^{ij} =\sum\limits_{bl} [\partial_a^i\partial_a^j \partial_b^l k(r)]_{aa b}^{ijl}Z_b^j.
  \label{eq:Ainf_H}
\end{equation}

\subsection{Dot Product Kernels}
\label{A:gH_dot}
\paragraph*{Gradient}
For dot product kernels the gradient at a point $\v{x}_a$ is readily available from Eq.~\eqref{eq:Ainf_g} and Eq.~\eqref{eq:Adotprod_def} as
\begin{equation*}
  \v{g}(\v{x}_a) = \Lambda Z(\v{k}'_{ab})\trans + \Lambda (X-\v{c})((\v{k}''_{ab})\trans \odot Z\trans (\v{x}_a-\v{c})). 
\end{equation*}
A prior mean for the gradient was omitted.
\paragraph*{Hessian}
The posterior mean of the Hessian in Eq.~\eqref{eq:Ainf_H} first requires the third derivative of the kernel. Differentiating Eq.~\eqref{eq:Adotprod_def} again yields
\begin{equation*}
\begin{split}
  {\partial_a}^i {\partial_a}^i {\partial_b}^l k(r) 
  &= k_{ab}''\cdot\Lambda^{jl}\cdot[\Lambda(\v{x}_b - \v{c} )]^i+k_{ab}''\cdot\Lambda^{il}\cdot[\Lambda(\v{x}_b - \v{c} )]^j + \delta_{ab}k''_{ab}\cdot \Lambda^{ij}\cdot [\Lambda(\v{x}_a - \v{c} )]^l \\
  &+k'''_{ab} [\Lambda(\v{x}_b - \v{c} )]^j [\Lambda(\v{x}_a - \v{c} )]^l [\Lambda(\v{x}_b - \v{c} )]^i
\end{split}
\end{equation*}
To perform the contraction in Eq.~\eqref{eq:Ainf_H} we first introduce $\tilde{X}=X-\v{c}$ and perform the contraction over $l$ which results in
\begin{equation*}
\begin{split}
  \bar{H}(\v{x}_a)^{ij} &=\sum\limits_b\, 
  k_{a b}''\cdot[\Lambda Z]_b^{j}\cdot[\Lambda\tilde{X}]_b^i + 
  k_{a b}''\cdot[\Lambda Z]_b^{i}\cdot[\Lambda\tilde{X}]_b^j + 
  \delta_{a b}\cdot  \Lambda^{ij}\cdot k''_{a b}\cdot[\Lambda(\v{x}_a - \v{c} )\trans\Lambda Z]_{a b} \\
  &+k'''_{a b}\cdot [\Lambda\tilde{X}]_b^i \cdot [\Lambda\tilde{X}]_b^j \cdot [(\v{x}_a - \v{c} )\trans\Lambda Z]_{a b}.
\end{split}
\end{equation*}
The final contraction of $b$ can easily be interpreted as standard matrix multiplication to arrive at the form
\begin{equation*}
\bar{H}(\v{x}_a)=
  \begin{bmatrix}
    \Lambda\tilde{X},\Lambda Z
  \end{bmatrix}
  \begin{bmatrix}
    M & \hat{M}\\
    \hat{M} & 0
  \end{bmatrix}
  \begin{bmatrix}
    \tilde{X}\trans\Lambda\\ 
    Z\trans \Lambda
  \end{bmatrix}
  + \Lambda \cdot \op{Tr}(\breve{M}).
\end{equation*}
All these $M$-matrices are diagonal matrices with $N$ elements 
\begin{align*}
M_{bb}
&=\v{k}_{a b}'''\odot[(\v{x}_a -\v{c})\trans\Lambda Z]_{a b}, \\
\hat{M}_{bb}
&=\v{k}''_{a b} \\
\breve{M}_{bb}
&=\delta_{a b} \cdot \v{k}''_{a b}(\v{x}_a-\v{c})\trans \Lambda Z. 
\end{align*}
The last expression including $\op{Tr}(\breve{M})$ can be simplified to $k''_{a a}(\v{x}_a-\v{c})\trans \Lambda Z$ \textbf{if} $\v{x}_a \in X$.

\subsection{Stationary Kernels}
\paragraph*{Gradient} inference for stationary kernels looks similar to the dot product kernels but has some important differences. For the following derivations we introduce $\tilde{k}'=2k'$, $\tilde{k}''=4k''$, $\tilde{k}'''=8k''$ and $\tilde{X}=(\v{x}_a - X)$.
The posterior mean gradient at a point $\v{x}_a$ for a stationary kernel is
\begin{equation}
\begin{split}
  \v{g}(\v{x}_a) &= -\Lambda Z\tilde{\v{k}}'_{ba} - \Lambda \tilde{X}(\tilde{\v{k}}''_{ba}\odot \v{m}_b),\\
  \v{m}_b &=(\sum\limits_l Z_b^l \odot [\Lambda \tilde{X}]_b^l)
\end{split}
\label{eq:Ainf_g_stat}
\end{equation}

\paragraph*{Hessian}
The third derivative of stationary kernels required for the Hessian inference is
\begin{equation*}
\begin{split}
  {\partial_a}^i {\partial_a}^i {\partial_b}^l k(r) 
  &= -\tilde{k}_{ab}''\cdot\Lambda^{jl}\cdot[\Lambda(\v{x}_a - \v{x}_b)]^i- \tilde{k}_{ab}''\cdot\Lambda^{il}\cdot[\Lambda(\v{x}_a - \v{x}_b)]^j 
  + \tilde{k}''_{ab}\cdot \Lambda^{ij}\cdot [\Lambda(\v{x}_a - \v{x}_b )]^l \\
  &-\tilde{k}'''_{ab} [\Lambda(\v{x}_a - \v{x}_b)]^j [\Lambda(\v{x}_a - \v{x}_b )]^l [\Lambda(\v{x}_a - \v{x}_b)]^i,
\end{split}
\end{equation*}
with $\v{m}_b$ the same vector as in Eq.~\eqref{eq:Ainf_g_stat}.
The posterior mean is obtained in the same way as for the dot product, by Eq.~\eqref{eq:Ainf_H}
\begin{equation*}
\begin{split}
  \bar{H}(\v{x}_a)^{ij} &= \sum\limits_b\, 
  -\tilde{k}_{a b}''\cdot[\Lambda Z]_b^{j}\cdot[\Lambda\tilde{X}]_b^i  
  -\tilde{k}_{a b}''\cdot[\Lambda Z]_b^{i}\cdot[\Lambda\tilde{X}]_b^j + 
  \Lambda^{ij}\cdot \tilde{k}_{a b}''\odot\v{m}_b \\
  &-(\tilde{k}_{a b}'''\odot \v{m}_b )\cdot[\Lambda\tilde{X}]_b^i \cdot [\Lambda\tilde{X}]_b^j.
\end{split}
\end{equation*}
The posterior mean can be written in standard matrix notation as
\begin{equation*}
\bar{H}(\v{x}_a)=
  \begin{bmatrix}
    \Lambda\tilde{X},\Lambda Z
  \end{bmatrix}
  \begin{bmatrix}
    M & \hat{M}\\
    \hat{M} & 0
  \end{bmatrix}
  \begin{bmatrix}
    \tilde{X}\trans\Lambda\\ 
    Z\trans \Lambda
  \end{bmatrix}
  + \Lambda \cdot \op{Tr}(\breve{M}).
\end{equation*}
The diagonal matrices are this time given by 
\begin{align*}
M_{bb}
&=\tilde{\v{k}}_{a b}'''\odot \v{m}_b,\\ 
\hat{M}
&=-\tilde{\v{k}}_{a b}'',\\
\breve{M}_{bb}
&=\tilde{\v{k}}_{a b}''\odot \v{m}_b.
\end{align*}

\section{Further Details about Applications} 

\subsection{Inferring the Optimizer}
\label{A:infer_min}

A GP with gradient observations learns a mapping $\v{x}\rightarrow \nabla f(\v{x})$. With efficient gradient inference we can also flip the inference and learn a mapping $\nabla f(\v{x}) \rightarrow \v{x}(\nabla f)$ and query what $\v{x}(\nabla f(\v{x})=0)$ for a new update.
This is achieved by performing gradient inference but interchanging the input and output.
The posterior mean for which $\v{x}$ $\nabla f(\v{x})=0$ occurs is
\begin{equation*}
  \bar{\v{x}}_* = \v{x}_m + [\nabla K \nabla'(0,G)]\left[ \nabla K \nabla'(G,G) \right]^{-1}\vect{X-\v{x}_m}.
\end{equation*}

\subsection{Stationary Linear Solvers}
\label{A:special_case}
For the special case of stationary linear solvers in linear algebra we have $f(\v{x})=\frac{1}{2}(\v{x}-\v{x}_*)\trans A (\v{x}-\v{x}_*)$  and $\nabla f(\v{x}) = \v{g}(\v{x}) =A (\v{x}-\v{x}_*)$ and we are interested in inferring $\v{x}_*$.

For the polynomial(2) kernel if we use $\v{c}=\v{g}_m$ and prior mean $\v{\mu}=\v{x}_m$ inference is fast.
First define $\tilde{X}=X-\v{x}_m$ and $\tilde{G}=\v{g}-\v{g}_m$.
Because $\tilde{G}\trans \tilde{X}=\tilde{X}\trans \tilde{G}$ we get the $Z$ that solves $\nabla K \nabla'\vect{Z}=\vect{\tilde{X}}$:
\begin{equation}
  Z = \Lambda^{-1}\tilde{X}(\tilde{G}\trans \Lambda \tilde{G})^{-1} - \frac{1}{2}\tilde{G}(\tilde{G}\trans\Lambda \tilde{G})^{-1}\tilde{G}\trans\tilde{X}(\tilde{G}\trans \Lambda \tilde{G})^{-1}
\end{equation}
Inferring at which the point $\hat{\v{x}}_a$ a gradient $\v{g}_a$ occurs is done by the following computation:
\begin{equation*}
  \begin{split}
   \hat{\v{x}}_a 
   &= \v{x}_m + \Lambda Z (\tilde{G}\trans \Lambda (\tilde{\v{g}}_a - \tilde{\v{g}}_m)) + \Lambda \tilde{X}[Z\trans \Lambda (\tilde{\v{g}}_a - \tilde{\v{g}}_m))]\\
   &=\v{x}_m 
    + \tilde{X}(\tilde{G}\trans \Lambda \tilde{G})^{-1}(\tilde{G}\trans \Lambda (\tilde{\v{g}}_a - \tilde{\v{g}}_m))
    - \frac{1}{2}\Lambda\tilde{G}(\tilde{G}\trans\Lambda \tilde{G})^{-1}\tilde{G}\trans\tilde{X}(\tilde{G}\trans \Lambda \tilde{G})^{-1} (\tilde{G}\trans \Lambda (\tilde{\v{g}}_a - \tilde{\v{g}}_m)) \\
    &+ \Lambda\tilde{G}[(\tilde{G}\trans \Lambda \tilde{G})^{-1}\tilde{X}\trans(\tilde{\v{g}}_a - \tilde{\v{g}}_m) 
    - \frac{1}{2} (\tilde{G}\trans \Lambda \tilde{G})^{-1} \tilde{G}\trans \tilde{X} (\tilde{G}\trans \Lambda \tilde{G})^{-1}\tilde{G}\trans \Lambda (\tilde{\v{g}}_a - \tilde{\v{g}}_m)
    ]\\
    &= \v{x}_m 
    + \tilde{X}(\tilde{G}\trans \Lambda \tilde{G})^{-1}\tilde{G}\trans \Lambda (\tilde{\v{g}}_a - \tilde{\v{g}}_m)\\
    &+ \Lambda\tilde{G}[(\tilde{G}\trans \Lambda \tilde{G})^{-1}\left( \tilde{X}\trans(\tilde{\v{g}}_a - \tilde{\v{g}}_m) 
    - \tilde{G}\trans \tilde{X} (\tilde{G}\trans \Lambda \tilde{G})^{-1}\tilde{G}\trans \Lambda (\tilde{\v{g}}_a - \tilde{\v{g}}_m)\right)]
  \end{split}
\end{equation*}

\section{Details about Experiments} 
\label{A:experiments}

\subsection{Linear Algebra}
\label{A:quad_opt}
For the linear algebra task we generated the matrix $A$ Eq.~\eqref{eq:Axb} in a manner beneficial for CG. 
The eigenvalues of $A$ were generated according to
\begin{equation*}
  \lambda_i = \lambda_{min} + \frac{\lambda_{max}-\lambda_{min}}{N-1}\cdot\rho^{N-i}\cdot(N-i),
\end{equation*}
with $\lambda_{min}=0.5$, $\lambda_{max}=100$ yielding a condition number of $\kappa(A)=200$ and $\rho=0.6$ so approximately the 15 largest eigenvalues are larger than 1. In this setting CG is expected to converge in slightly more than 15 iterations. A relative tolerance in gradient norm of $10^{-5}$ was used as termination criterion due to numerical instabilities. The starting and solution points were sampled according to $\v{x}_0\sim \mathcal{N}(0,5^2\cdot I)$ and $\v{x}_*\sim \mathcal{N}(-2\cdot \v{1},I)$. 
The Hessian-based optimization used a fixed $\v{c}=\v{0}$ and $\v{g}_c=A(\v{c}-\v{x}_*)=-A\v{x}_*=-\v{b}$ in the linear system interpretation $A\v{x}=\v{b}$. There a plenty of possibilities for how the algorithm can be implemented and this particular version was sensitive to the relative position of $\v{c}$ and $\v{x}_*$.

\subsection{Nonlinear Optimization}
\label{A:nonlin_opt}
We chose the test function (restated here for convenience)
\begin{equation*}
  f(\v{x})=\sum_{i=1}^{D-1} x_i^2 + 2\cdot(x_{i+1}-x_i^2)^2
\end{equation*}
for the more challenging nonlinear experiments. It is a relaxed version of the famous Rosenbrock function, which was used to better control the magnitude of the gradients for the high-dimensional problem. This was important because the RBF kernels used for the optimization used a fixed $\Lambda$, which could lead to numerical issues if the magnitude of the steps and gradients drastically changed between iterations. 
The lengthscale of the isotropic kernels in the algorithms were $\Lambda = 9\cdot I$ for GP-H and $\Lambda=0.05\cdot I$. There are too many options of extending the algorithm to go over in this manuscript, which is why the algorithm should be seen more as a proof-of-concept than radical new algorithm.

\subsection{Hamiltonian Monte Carlo}
\label{A:hmc}

We used the following unnormalized density as a target for the \hmc~experiment
\begin{equation}
 f(x) = \exp \left(- \frac{1}{2} \left(x_1^2 + (a_0 x_1^2 + a_1 x_2 + a_2)^2 + \sum_{i=3}^D a_i x_i^2 \right) \right)
\label{eq:banana}
\end{equation}
and set the parameter vector to $\v{a} = [2, -2, 2, \dots, 2]\trans$.
The distribution is thus Gaussian with variance $\frac{1}{2}$ in all components other than $x_1$ and $x_2$.
Since we use an isotropic RBF kernel to model the potential energy (i.e.,~the negative logarithm of the above function), we randomly rotate the above function by applying sampled orthonormal matrices to the input vector.

Fig.~\ref{fig:banana} uses Eq.~\eqref{eq:banana} directly, and thus the kernel is aligned with the problem.
We choose a (squared) lengthscale of $0.4 D$ where $D = 100$ from visual inspection of the typical scale of the ``banana''.
\hmc~uses a step-size $\epsilon = 4\cdot 10^{-3} / \lceil\sqrt[4]{D}\rceil$ and number of leapfrog steps $T = 32 \cdot \lceil\sqrt[4]{D}\rceil$, with the term $\lceil\sqrt[4]{D}\rceil$ being motivated by the analysis of how these parameters should change with increasing dimension \citep{neal2012mcmc}.
For all experiments we draw a standard normal vector as a starting point and simulate $D$ times with plain \hmc~for burn-in, before retaining samples in the case of \hmc, or starting the training procedure for \gpghmc.
The training is performed as described in Sec.~\ref{sub:hmc_exp}.

The rotated version of the above function used slightly different parameters for the RBF kernel, a squared lengthscale of $0.25 D$ to stay on the conservative side about the target function.
Also we halved the stepsize of the leapfrog integrator while leaving the number of steps taken unchanged.
Otherwise, the acceptance rate also dropped significantly for both methods.
All experiments used a mass parameter of $m=1$.

Algorithm \ref{alg:gpg-hmc} summarizes the \gpghmc~method without the training procedure which leaves a lot of space for engineering.
In fact, this is identical to standard \hmc, except for the fact that instead of the true gradient $\nabla E$ the \gp~surrogate $\widehat{\nabla E}$ is used.

\begin{algorithm}[h]
   \caption{GPG-HMC}
   \label{alg:gpg-hmc}
\begin{algorithmic}
	\INPUT $\v{x}_0$, $E(\cdot)$, $\widehat{\nabla E}(\cdot)$, $N$, $T$, $\epsilon$, $m$
	\OUTPUT $\v{X}$
	\STATE $\v{x} = \v{x}_0$; $\v{X} = [\,]$
	\FOR{n = 1:N}{
	\STATE $\v{p} \sim \mathcal{N} (0, mI)$
	\STATE $H \gets U(\v{x}) + \frac{\v{p}\Trans \v{p}}{2m}$
	\STATE $\v{x}_{\rm new}, \v{p} \gets \text{\textsc{Leapfrog}} (\v{x}, \v{p}, \widehat{\nabla E}(\cdot), T, \epsilon)$
	\STATE $\Delta H \gets E(\v{x}) + \frac{\v{p}\Trans \v{p}}{2m} - H$
	\IF{$r\sim\text{\textsc{Uniform}}[0,1] < \min (1, \operatorname{e}^{-\Delta H})$} 
	\STATE $\v{x} \gets \v{x}_{\rm new}$
	\ENDIF
	\STATE $\v{X} \gets [\v{X}, \v{x}]$
	}
	\ENDFOR
\end{algorithmic}
\end{algorithm}

\end{document}